\documentclass{article}
\usepackage{ijcai23}

\usepackage{times}
\usepackage{soul}
\usepackage{url}
\usepackage[hidelinks]{hyperref}
\usepackage[utf8]{inputenc}
\usepackage[small]{caption}
\usepackage{graphicx}
\usepackage{amsmath}
\usepackage{amsthm}
\usepackage{booktabs}
\usepackage{algorithm}
\usepackage{algorithmic}
\usepackage[switch]{lineno}
\usepackage{bbding}
\usepackage{caption}

\title{Q-Cogni: An Integrated Causal Reinforcement Learning Framework}

\author{
Cristiano da Costa Cunha$^1$
\and
Wei Liu$^1$\and
Tim French$^1$\and
Ajmal Mian$^1$
\affiliations
$^1$Department of Computer Science, University of Western Australia\\
\emails
  cris.dacostacunha@research.uwa.edu.au,
\{wei.liu, tim.french, ajmal.mian\}@uwa.edu.au
}

\begin{document}

\maketitle

\begin{abstract}
We present \textit{Q-Cogni}, an algorithmically integrated causal reinforcement learning framework that redesigns \textit{Q-Learning} with an autonomous causal structure discovery method to improve the learning process with causal inference. \textit{Q-Cogni} achieves optimal learning with a pre-learned structural causal model of the environment that can be queried during the learning process to infer cause-and-effect relationships embedded in a state-action space. We leverage on the sample efficient techniques of reinforcement learning, enable reasoning about a broader set of policies and bring higher degrees of interpretability to decisions made by the reinforcement learning agent. We apply \textit{Q-Cogni} on the Vehicle Routing Problem (VRP) and compare against state-of-the-art reinforcement learning algorithms. We report results that demonstrate better policies, improved learning efficiency and superior interpretability of the agent's decision making. We also compare this approach with traditional shortest-path search algorithms and demonstrate the benefits of our causal reinforcement learning framework to high dimensional problems. Finally, we apply \textit{Q-Cogni} to derive optimal routing decisions for taxis in New York City using the Taxi \& Limousine Commission trip record data and compare with shortest-path search, reporting results that show 85\% of the cases with an equal or better policy derived from \textit{Q-Cogni} in a real-world domain.
\end{abstract}

\section{Introduction}
Evidence suggests that the human brain operates as a dual system. One system learns to repeat actions that lead to a reward, analogous to a model-free agent in reinforcement learning, the other learns a model of the environment which is used to plan actions, analogous to a model-based agent. These systems coexist in both cooperation and competition which allows the human brain to negotiate a balance between cognitively cheap but inaccurate model-free algorithms and relatively precise but expensive model-based algorithms \cite{Gershman2017}.

Despite the benefits causal inference can bring to autonomous learning agents, the degree of integration in artificial intelligence research are limited. This limitation becomes a risk, in particular that data-driven models are often used to infer causal effects. Solely relying on data that is never bias-free, eventually leads to untrustworthy decisions and sub-optimal interventions \cite{Prosperi2020}. 

In this paper, we present \textit{Q-Cogni} a framework that integrates autonomous causal structure discovery and causal inference into a model-free reinforcement learning method. There are several emergent methods for integrating causality with reinforcement learning such as reward correction \cite{Buesing2018}, meta-reinforcement learning \cite{dasgupta2019causal}, latent causal-transition models \cite{gasse2021causal}, schema networks \cite{kansky2017schema} and explainable agents \cite{Madumal2020}. However, no method presents an approach that embeds causal reasoning, from an autonomously derived causal structure of the environment, during the learning process of a reinforcement learning agent to guide the generation of an optimal policy. Thus, our approach is able to target improvements in policy quality, learning efficiency and interpretability concurrently. 

\textit{Q-Cogni} samples data from an environment to discover the causal structure describing the relationship between state transitions, actions and rewards. This causal structure is then used to construct a \textit{Bayesian Network} which is used during a redesigned \textit{Q-Learning} process where the agent interacts with the environment guided by the probability of achieving the goal and receiving rewards in a probabilistic manner. The causal structure integrated with the learning procedure delivers higher sample efficiency as it causally manages the trade-off between exploration and exploitation, is able to derive a broader set of policies as rewards are much less sparse and provides interpretability of the agent's decision making in the form of conditional probabilities related to each state transition for a given set of actions.

 We validate our approach on the Vehicle Routing Problem (VRP) \cite{toth2002overview}. We start by comparing optimal learning metrics against state-of-the-art reinforcement learning algorithms \textit{PPO} \cite{schulman2017proximal} and \textit{DDQN} \cite{van2016deep}, using the \textit{Taxi-v3} environment from \textit{OpenAI Gym} \cite{brockman2016openai}. We also compare the advantages and disadvantages of \textit{Q-Cogni} against the shortest-path search algorithms \textit{Djikstra's} \cite{dijkstra1959note} and \textit{A*} \cite{Hart1968} with a particular focus in understanding applicability and scalability. Finally, we run experiments in a real-world scale problem, using the New York City TLC trip record data, which contains all taxi movements in New York City from 2013 to date \cite{NYC_data}, to validate \textit{Q-Cogni's} capabilities to autonomously route taxis for a given pickup and drop-off.

Our contributions with \textit{Q-Cogni} are three-fold. Firstly, \textit{Q-Cogni} is the first fully integrated, explainable, domain-agnostic and hybrid model-based and model-free reinforcement learning method that introduces autonomous causal structure discovery to derive an efficient model of the environment and uses that causal structure within the learning process. Secondly, we redesigned the \textit{Q-Learning} algorithm to use causal inference in the action selection process and a probabilistic Q-function during training in order to optimise policy learning. Finally, through extensive experiments, we demonstrate \textit{Q-Cogni's} superior capability in achieving better policies, improved learning efficiency and interpretability as well as near-linear scalability to higher dimension problems in a real-world navigation context.

\section{Background}
The focus of this work lies at the unification of causal inference and reinforcement learning. This is an emerging field that aims to overcome challenges in reinforcement learning such as 1) the lack of ability to identify or react to novel circumstances agents have not been programmed for \cite{darwiche2018human,chen2018lifelong}, 2) low levels of interpretability that erodes user's trust and does not promote ethical and unbiased systems \cite{ribeiro2016should,marcus2018deep} and 3) the lack of understanding of cause-and-effect relationships \cite{Pearl2010}.

Our approach builds upon a wealth of previous contributions to these areas, which we briefly cover below.

\vspace{1mm}
\noindent {\bf Causal Structure Discovery.} Revealing causal information by analysing observational data, i.e. ``causal structure discovery", has been a significant area of recent research to overcome the challenges with time, resources and costs by designing and running experiments \cite{Kuang2020}. 

Most of the work associated with integrating causal structure discovery and reinforcement learning have been focused on using reinforcement learning to discover cause-and-effect relationships in environments which agents interact with to learn \cite{zhu2019causal,wang2021ordering,huang2020causal,amirinezhad2022active,sauter2022meta}. To our knowledge, a small amount of work has explored the reverse application, such as schema networks \cite{kansky2017schema}, counterfactual learning \cite{lu2020sample} and causal MDPs \cite{lu2022efficient}. 

We build upon this work and redesign the way in which structure causal models (SCMs) are used. In the related work they are typically used to augment input data with what-if scenarios a-priori to the agent learning process. In our approach, the SCM is embedded as part of a redesigned Q-Learning algorithm and only used during the learning process. Our approach also enables learning a broader set of policies since what-if scenarios are estimated for each state-action pair during the learning process. This not only improves policy optimality but also
provides a superior sample efficiency as it allows for “shortcutting” the exploration step during the learning process.  

\vspace{1mm}
\noindent {\bf Causal Inference.} Recent work has demonstrated the benefits of integrating causal inference in reinforcement learning. 

In Seitzer, Sch\"{o}lkopf, and Martius \shortcite{NEURIPS2021_c1722a79} the authors demonstrate improvement in policy quality by deriving a measure that captures the causal influence of actions on the environment in a robotics control environment and devise a practical method to integrate in the exploration and learning of reinforcement learning agents.

In Yang et al. \shortcite{yang2022training} the authors propose an augmented DQN algorithm which receives interference labels during training as an intervention into the environment and embed a latent state into its model, creating resilience by learning to handle abnormal event (e.g. frozen screens in Atari games).

In Gasse et al. \shortcite{gasse2021causal} the authors derive a framework to use a structural causal model as a Partially Observable Markov Decision Process (POMDP) in model-based reinforcement learning.

Leveraging upon these concepts, in our approach we expand further the structural causal model and fit it with a \textit{Bayesian Network}. This enables our redesigned Q-Learning procedure to receive rewards as a function of the probability of achieving a goal for a given state-action pair, significantly improving the sample efficiency of the agent, as in each step the agent is able to concurrently derive dense rewards for several state transitions regulated by the causal structure. To our knowledge this is an integration perspective not yet explored.

\vspace{1mm}
\noindent {\bf Model-Free Reinforcement Learning.}
Centre to the reinforcement learning paradigm is the learning agent, which is the ``actor" that learns the optimal sequence of actions for a given task, i.e. the optimal policy. As this policy is not known a priori, the aim is to develop an agent capable of learning it by interacting with the environment \cite{kaelbling1996reinforcement}, an approach known as model-free reinforcement learning.

Model-free reinforcement learning relies on algorithms that sample from experience and estimate a utility function such as \textit{SARSA}, \textit{Q-Learning} and \textit{Actor-Critic} methods \cite{arulkumaran2017brief}. Recent advances in deep learning have promoted growth of model-free methods\cite{ccalicsir2019model}. However, whilst model-free reinforcement learning is a promising area to enable human-level artificial intelligence, it comes with its own limitations. These are the applicability restricted to a narrow set of assumptions (e.g. a Markov Decision Process) that is not necessarily reflective of the dynamics of the real-world environment \cite{StJohn2020}; lower performance when evaluating off-policy decisions (i.e. policies different than those contained in the underlying data used by the agent) \cite{Bannon2020} and perpetual partial observability since sensory data provide imperfect information about the environment and hidden variables are often the ones causally related to rewards \cite{Gershman2017}. These limitations are a disadvantage of model-free reinforcement learning which can be overcome with explicit models of the environment, i.e. model-based reinforcement learning. However, whilst the model-based approach would enhance sample efficiency, it also would come at a cost of increased computational complexity as many more samples are required to derive an accurate model of the environment \cite{polydoros2017survey}. 

In our work we use causal structure discovery to simultaneously provide model-free reinforcement learning agents with the ability of dealing with imperfect environments (e.g., latent variables) and maintain sample efficiency of a model-based approach. A hybrid approach not extensively explored to our knowledge.

\section{Q-Cogni}
\begin{figure}[t]
\centering
\includegraphics[width=1\columnwidth, height=300pt]{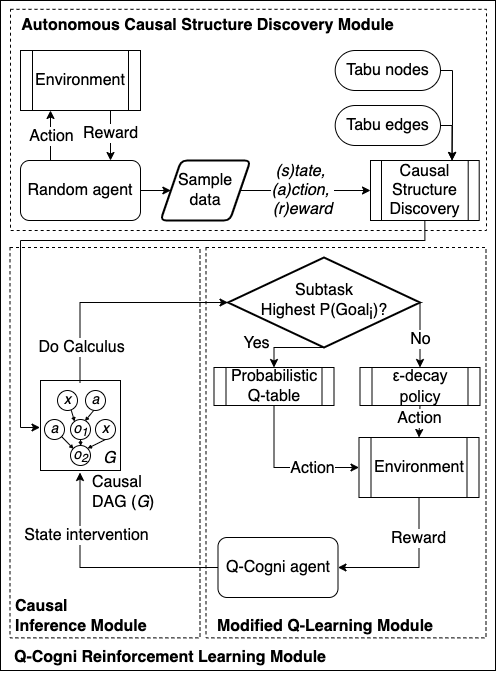}
\caption{Q-Cogni causal reinforcement learning framework with its modules highlighted.}
\label{CRLframework}
\end{figure}

We present \textit{Q-Cogni}, a framework that integrates autonomous causal structure discovery, causal inference and reinforcement learning. Figure \ref{CRLframework} illustrates the modules and interfaces which we further detail below.

\subsection{Autonomous Causal Structure Discovery}
The first module in \textit{Q-Cogni} is designed to autonomously discover the causal structure contained in an environment. It starts by applying a random walk in the environment while storing the state, actions and rewards. The number of steps required to visit every state in the environment with a random walk is proportional to the harmonic number, approximating the natural logarithm function which grows without limit, albeit slowly, demonstrating the efficiency of the process. This sampled dataset contains all the information necessary to describe the full state-action space and its associated transitions. A further benefit of our approach is that this step only needs to be performed once regardless of the environment configuration.

We use the \textit{NOTEARS} algorithm \cite{Zheng2018} in the \textit{Q-Cogni} framework, providing an efficient method to derive the causal structure encoded in the dataset sampled from the environment.

The resulting structure learned is then encoded as a DAG \textit{G} with nodes \textit{v} $\in$ \textit{G}, state variables \textit{x}, actions \textit{a} and edges \textit{e} $\in$ \textit{G} which represent the state transition probabilities. With a maximum likelihood estimation procedure, the discovered structure is then fitted with the dataset sample generated to estimate the conditional probability distributions of the graph and encode it as a \textit{Bayesian Network}. 

Whilst this module focuses on autonomous causal structure learning, \textit{Q-Cogni} provides the flexibility to receive human inputs in the form of tabu nodes and edges, i.e. constraints in which a human expert can introduce in the causal structure discovery procedure. This capability allows integration between domain knowledge with a data-driven approach providing a superior model in comparison to using either in isolation.

\subsection{Causal Inference}
We leverage upon the causal structure model discovered and the \textit{Bayesian Network} of the environment estimated to provide \textit{Q-Cogni's} reinforcement learning module with causal inference capabilities. 

The causal inference sub-module uses the causal DAG $G(V, E)$ and receives from the \textit{Q-Cogni} agent a single state $s \in S$ containing each state variable $x \in s$ with values sampled from the environment $M$, the actions list $A$ containing each action $a \in A$ and the first-priority sub-goal $o$ to be solved. The marginals in each node $v \in V \cap s$ are updated with the state variable values $x$ $\forall ~x$ $\in s$. The procedure described in Algorithm \ref{alg:q-cognicausalinference} selects the best action $a^*$ and calculates the associated $P(o={\rm True}|x, A)$ where $a^* \in A$. This is analogous to a probabilistic reward estimation \textit{r} for the given $(s, a^*)$ pair.

\begin{algorithm}[tb]
\begin{flushleft}
\caption{Q-Cogni causal inference routine}
\label{alg:q-cognicausalinference}
\textbf{Procedure}: INFER-MAX-PROB(G, s, A, o) \\
\textbf{Input}: $G(V)$: causal structure DAG as a function of nodes $v \in V$ where $V = \{s, A, O\}$ with a fitted $Bayesian Network$ containing $P(v|$ \textit{parents of} $v)$ $\forall v \in V$, $s$: a single state $s \in S$ containing state variables $x \in s$, $A$: actions list with each action $a \in A$ where $a$ values $\in \{True, False\}$ , $o$: node representing goal to be solved where $o \in O$ \\
\textbf{Output}: $a^*$: action $a \in A$ where $a = True$ $\land A \setminus a = False $ for $\max p$, $p$: $P(o = True|x, A)$ where $a^* \in A$ \\
\begin{algorithmic}[1]

\STATE Let $p = 0$, $a = False$ $\forall a \in A$, $a^* = $ \O 

\FOR{each $v \in V \cap s$}
    \STATE * Update each node $v$ representing a state variable $x \in s$ with its value
    \STATE $v \gets x$
\ENDFOR

\FOR{each $a \in A$}
    \STATE * Calculate the probability of $g = True$ for each action $a \in A$ when $a = True$
    \STATE $a \gets True$
    \STATE $a^- \gets False$ $\forall a^- \in A \setminus a$ 
    \IF{$p < P(o = True| V \cap s, A)$}
    \STATE $p \gets P(o = True| V \cap s, A)$ 
    \STATE $a^* \gets a$ 
    \ENDIF
\ENDFOR

\STATE \textbf{return} $a^*$, $p$
\end{algorithmic}
\end{flushleft}
\end{algorithm}

This module enables the gains in learning efficiency by the agent as it shortcuts the reinforcement learning exploration procedure through the structural prior knowledge of the conditional probability distributions of $(s,a)$ pairs encoded in the DAG. It also provides explicit interpretability capabilities to the \textit{Q-Cogni} reinforcement learning module by being able to estimate $P(o={\rm True}|x, A)$. 

\subsection{Modified Q-Learning}
The modified \textit{Q-Learning} module uses a hybrid learning procedure that uses \textit{Q-Cogni's} causal inference module and a $\epsilon$-decay exploration strategy. In addition, one central idea in \textit{Q-Cogni} is to use the inherent and known structure of the reinforcement learning sub-goals for a given task to reduce the problem dimensionality. Whilst this is not a strict limitation to the approach, when available a-priori it gives a significant advantage in computational efficiency for the learning procedure as \textit{Q-Cogni} shrinks the $state-action$ space to the subset that matters for a given sub-goal. \textit{Q-Cogni's} learning procedure can receive a prioritised list $O$ of ordered reinforcement learning sub-goals $o$ and uses that information to choose when to ``explore vs. infer". If such a goal sequence is not known a-priori the benefits from our approach still hold, albeit a lower sample efficiency but still superior to a traditional agent that would require balancing exploration vs. exploitation. 

To achieve that, for the prioritised sub-goal $o$, \textit{Q-Cogni} assesses if $ \max P(o=True|x, A)$ takes place when $a^* \in A$ is a $parent$ node $\in V$ of the sub-goal node $o$. In this case, the agent selects $a^*$ and directly applies into the environment to obtain the reward \textit{r} adjusted by $P(o=True|x, A)$ during the value function update procedure, a step taken to avoid reward sparsity and improve learning performannce. The Q-table stores this result, providing a more robust estimation of value without having to perform wide exploration in the environment in contrast to unadjusted rewards. Otherwise, the \textit{Q-Cogni} agent will perform the $\epsilon$-decay exploration procedure. Algorithm \ref{alg:q-cognimodifiedq-learning} describes the modified \textit{Q-Learning} routine in \textit{Q-Cogni}.

\begin{algorithm}[tb]
\begin{flushleft}
\caption{Q-Cogni modified Q-Learning}
\label{alg:q-cognimodifiedq-learning}
\textbf{Procedure}: Q-COGNI-LEARN(G, M, A, O)\\
\textbf{Input}: $G(V)$: causal structure DAG as a function of nodes $v \in V$ and $V = \{s, A, O\}$, $M$: environment containing a list $S$ of states $s \in S$ where $s = \{s_{i} \dots s_{t}\}$, $A$: actions containing list of actions $a \in A$, $O$: sequence of goal nodes $o_j \in O \in V$ and $j \in [1, \dots, n_{goals}]$ in a priority order \\
\textbf{Parameters}: $N$: number of episodes, $\alpha$: learning rate, $\gamma$: discount rate, $\epsilon$: initial threshold for exploration, $\epsilon_{min}$: minimum $\epsilon$, $\delta$: decay rate for $\epsilon$ \\
\textbf{Output}: Q-table with $Q(s,a)$ pairs estimating the optimal policy $\pi$*
\begin{algorithmic}[1] 
\STATE Initialise $Q(s,a)$ arbitrarily;
\FOR{each episode $n \in [1, \dots, N]$}
\STATE Initialise state $s$ = $s_i$ from environment $M$;
\STATE Let $j = 1$, $a = \emptyset $
    \WHILE{$s$ $\ne$ $s_t$}
            \STATE $a^*, p = $ INFER-MAX-PROB(G, s, A, $o_j$)
            \IF {$a^* \in \textit{parents of}$ $o_j$}
                \STATE $a \gets a^*$
                \STATE $j \gets j + 1$
            \ELSE 
                \STATE $\mu \gets u \sim \mathcal{U}(0,1)$
                \IF{$\mu < \epsilon $}
                    \STATE $a \gets $ RANDOM($A$)
                \ELSE
                    \STATE $a \gets \max Q(s,.)$
                \ENDIF
                \STATE $\epsilon \gets \max(\epsilon_{min}, \epsilon*\delta)$
            \ENDIF 
            \STATE * Apply action $a$ in environment $M$ with state $s$, observe reward $r$ and next state $s'$
            \STATE $Q(s, a) \gets Q(s, a) + \alpha \cdot (r \cdot p + \gamma \cdot \max_{a} Q(s', a) - Q(s, a))$
            \STATE $s \gets s'$
    \ENDWHILE 
\ENDFOR
\STATE \textbf{return} $Q(s, a)$, $\pi$*
\end{algorithmic}
\end{flushleft}
\end{algorithm}

This routine enables optimised learning. Policies are improved by the upfront knowledge acquired with the causal structure module and the representation of the state transition outcomes. Unnecessary exploration of state-action pairs that do not improve the probability of achieving the learning goal are eliminated, thus improving learning efficiency. 

\section{Approach Validation}
To validate our approach, we start with
the Vehicle Routing Problem (VRP). 
Here, we formally define the VRP problem and briefly discuss traditional solutions on which we build ours upon.

\vspace{1mm}
\noindent {\bf VRP.}
In our work, inspired by the emergence of self-driving cars, we use a variant of the VRP, where goods need to be picked up from a certain location and dropped off at their destination. The pick-up and drop-off must be done by the same vehicle, which is why the pick-up location and drop-off location must be included in the same route \cite{braekers2016vehicle}. 

This variant is known as the VRP with Pickup and Delivery, a NP-hard problem extensively studied by the operations research community given its importance to the logistics industry. The objective is to find the least-cost tour, i.e. the shortest route, to fulfill the pickup and drop-off requirements \cite{ballesteros2016review}.  

\vspace{1mm}
\noindent {\bf Shortest-Path Search Methods.}
The shortest-path search problem is one of the most fundamental problems in combinatorial optimisation. As a minimum, to solve most combinatorial optimisation problems either shortest-path search computations are called as part of the solving procedure or concepts from the framework are used \cite{gallo1986shortest}. Similarly, it is natural to solve the VRP with Pickup and Delivery with shortest-path search methods.

Despite successful shortest-path search algorithms such as Djikstra's and A*, VRP as a NP-hard problem can be very challenging for these methods. Exact algorithms like Djikstra's can be computationally intractable depending on the scale of the problem \cite{drori2020learning}; approximate algorithms like A* provide only worst-case guarantees and are not scalable \cite{williamson2011design}. Reinforcement learning is an appealing direction to such problems as it provides a generalisable, sample efficient and heuristic-free method to overcome the characteristic computational intractability of NP-hard problems.

\section{Experimental Results}
We start with the \textit{Taxi-v3} environment from \textit{OpenAI Gym} \cite{brockman2016openai}, a software implementation of an instance of the VRP with Pickup and Delivery. The environment was first introduced by Dietterich \shortcite{dietterich2000hierarchical} to illustrate challenges in hierarchical reinforcement learning.

\begin{figure}[t]
\centering
\includegraphics[width=1\columnwidth, trim=4 4 4 4,clip]{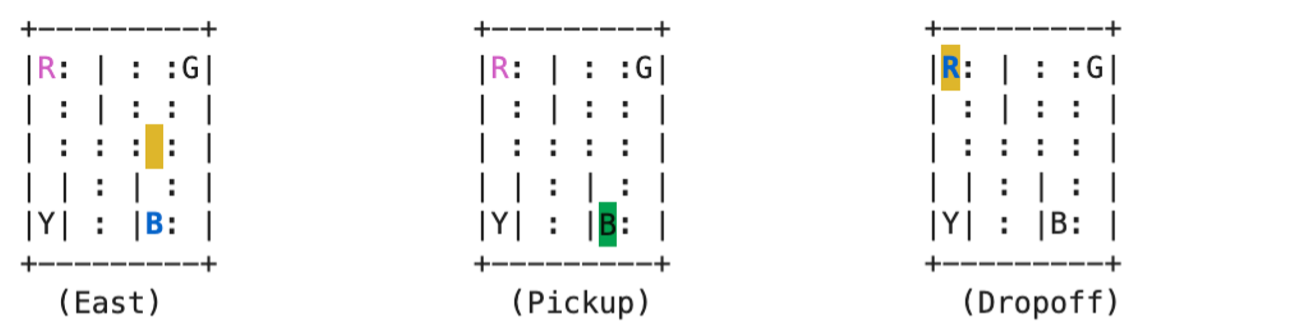}
\caption{\textit{Taxi-v3} environment showing a sequence of actions derived by a \textit{Q-Learning} agent.}
\label{QLearningTaxiv3}
\end{figure}

Figure \ref{QLearningTaxiv3} illustrates the \textit{Taxi-v3} environment with the example of a solution using the \textit{Q-Learning} algorithm. The 5$\times$5 grid has four possible initial locations for the passenger and destination indicated by R(ed), G(reen), Y(ellow), and B(lue). The objective is to pick up the passenger at one location and drop them off in another. The agent receive a reward of +20 points for a successful drop-off, and receives a reward of -1 for every movement. There are 404 reachable discrete states and six possible actions by the agent (move west, move east, move north, move south, pickup and deliver).

All experiments were performed on a p4d.24xlarge GPU enabled AWS EC2 instance.

\subsection{Optimal Learning}
We trained \textit{Q-Cogni}, \textit{Q-Learning}, \textit{DDQN} and \textit{PPO} algorithms for 1,000 episodes in the \textit{Taxi-v3} environment. \textit{DDQN} and \textit{PPO} were implemented using the \textit{Rlib} python package \cite{liang2018rllib}. Hyperparameters for \textit{DDQN} and \textit{PPO} were tuned using the \textit{BayesOptSearch} module, using 100 trials over 1,000 episodes each.

\subsubsection{Results and Discussion.}
\begin{figure}[t]
\centering
\includegraphics[width=1\columnwidth, height=160pt, trim=4 4 4 4,clip]{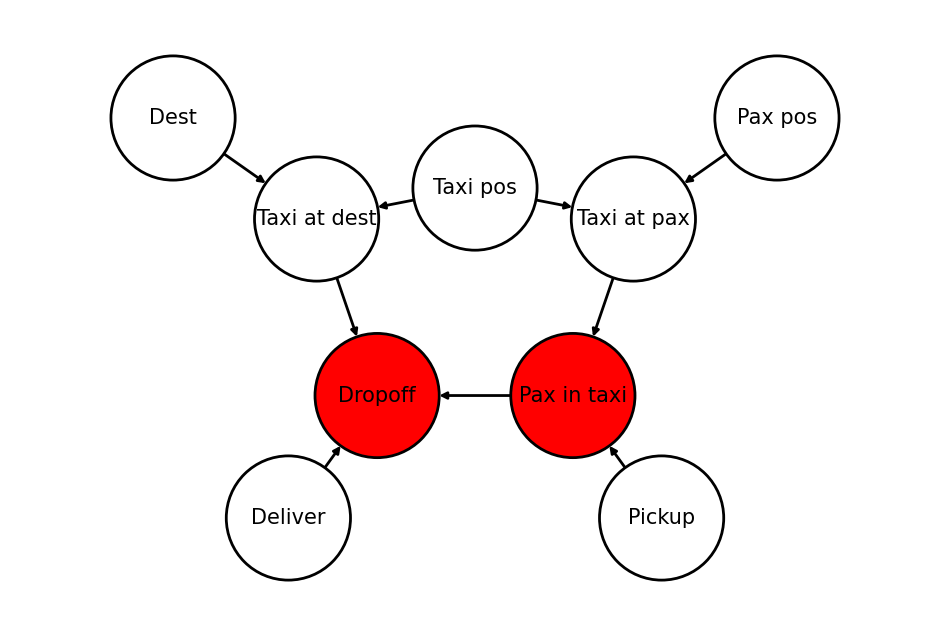}
\caption{Discovered causal structure for the \textit{Taxiv3} environment. Sub-goal nodes are highlighted in red.}
\label{SCMTaxi}
\end{figure}

Figure \ref{SCMTaxi} illustrates the autonomously discovered structure for the \textit{Taxiv3} environment using \textit{Q-Cogni}, after 500,000 samples collected with a random walk. We used the implementation of the \textit{NOTEARS} algorithm in the \textit{CausalNex} python package \cite{Beaumont_CausalNex_2021} to construct the causal structure model and fit the conditional probability distributions through a \textit{Bayesian Network}. The relationships discovered are quite intuitive demonstrating the high performance of the method. For example, for the node \textit{passenger in taxi} to be $True$ the nodes \textit{taxi on passenger location} and \textit{pickup action} must be $True$. 

The only domain related inputs given to the algorithm were constraints such as the sub-goal node 1 \textit{pax in taxi} cannot be a child node of the sub-goal 2 node \textit{drop-off} and location nodes must be a parent node. In addition, the list of ordered sub-goals was provided to \textit{Q-Cogni's} reinforcement learning module as [\textit{pax in taxi, drop-off}].

\begin{figure}[t]
\centering
\includegraphics[width=1\columnwidth,height=4.5cm, trim=4 4 4 4,clip]{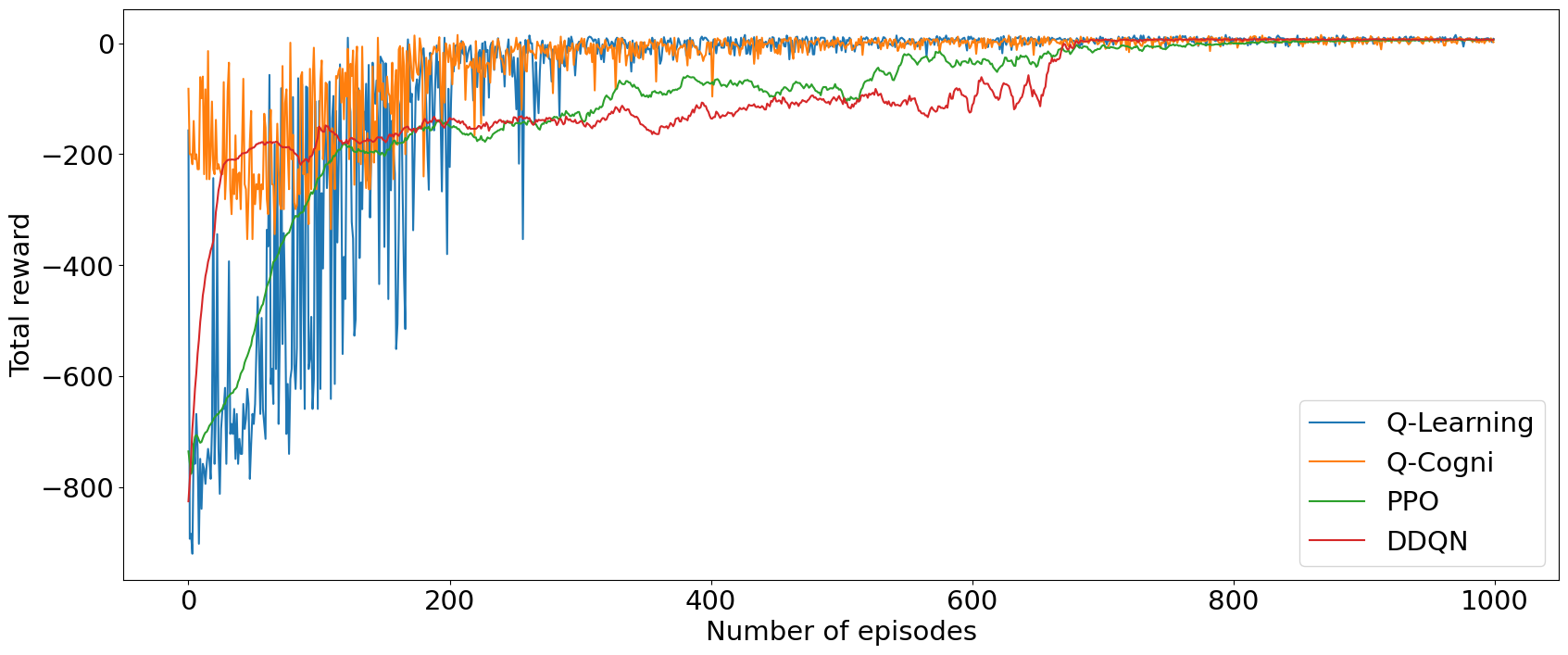}
\caption{Total reward vs. number of episodes for comparative reinforcement learning methods.}
\label{Q-CognivsRL}
\end{figure}

\begin{figure}[t]
\centering
\includegraphics[width=1\columnwidth, trim=4 4 4 4,clip]{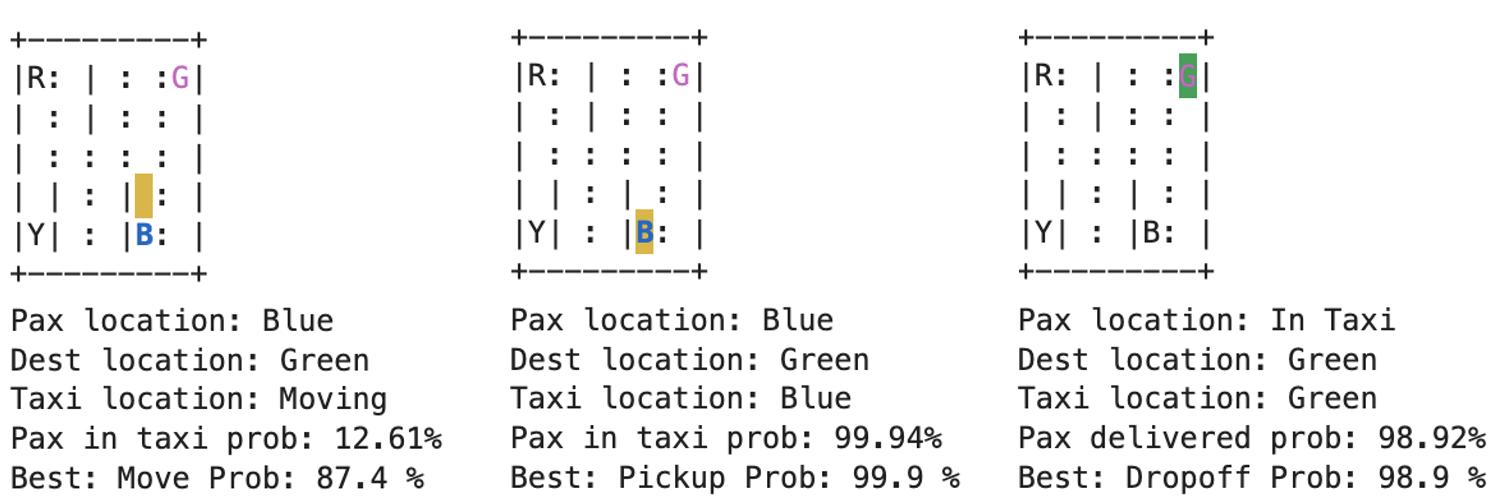}
\caption{\textit{Q-Cogni} interpretability. Sequence of decisions in a randomly selected episode post training.}
\label{Interpretability}
\end{figure}

Figure \ref{Q-CognivsRL} shows the results achieved over the 1,000 training episodes. We observe that all methods present similar policy performance (total reward per episode towards the end of training). However, \textit{Q-Cogni} achieves superior stability and learning efficiency in comparison to all other methods, as it is able to use the causal structure model and its causal inference capability to accelerate the action selection process when interacting with the environment. In addition, Figure \ref{Interpretability} demonstrates the interpretability capabilities of \textit{Q-Cogni}. At each step, a probability of the best action to be taken is provided allowing for better diagnostics, tuning and most importantly assessment of possible biases built into autonomous agents such as \textit{Q-Cogni}.

\subsection{Comparison to Shortest-Path Search Methods}
 We  analyse the characteristics of our approach against shortest-path search methods to highlight the advantages and disadvantages of  \textit{Q-Cogni}. We perform experiments in which expand the \textit{Taxi-v3} environment into larger state sizes, represented by a grid of \textit{n$\times$m} rows and columns. We then compare the time taken to achieve an optimal tour against Djikstra's algorithm and A$^*$ using a Manhattan distance heuristic.

\subsubsection{Results and Discussion.}
We report our comparison analysis across dimensionality, prior knowledge requirements, transportability between configurations and interpretability.

\begin{figure}[t]
\centering
\includegraphics[width=1\columnwidth, trim=4 4 4 4, clip]{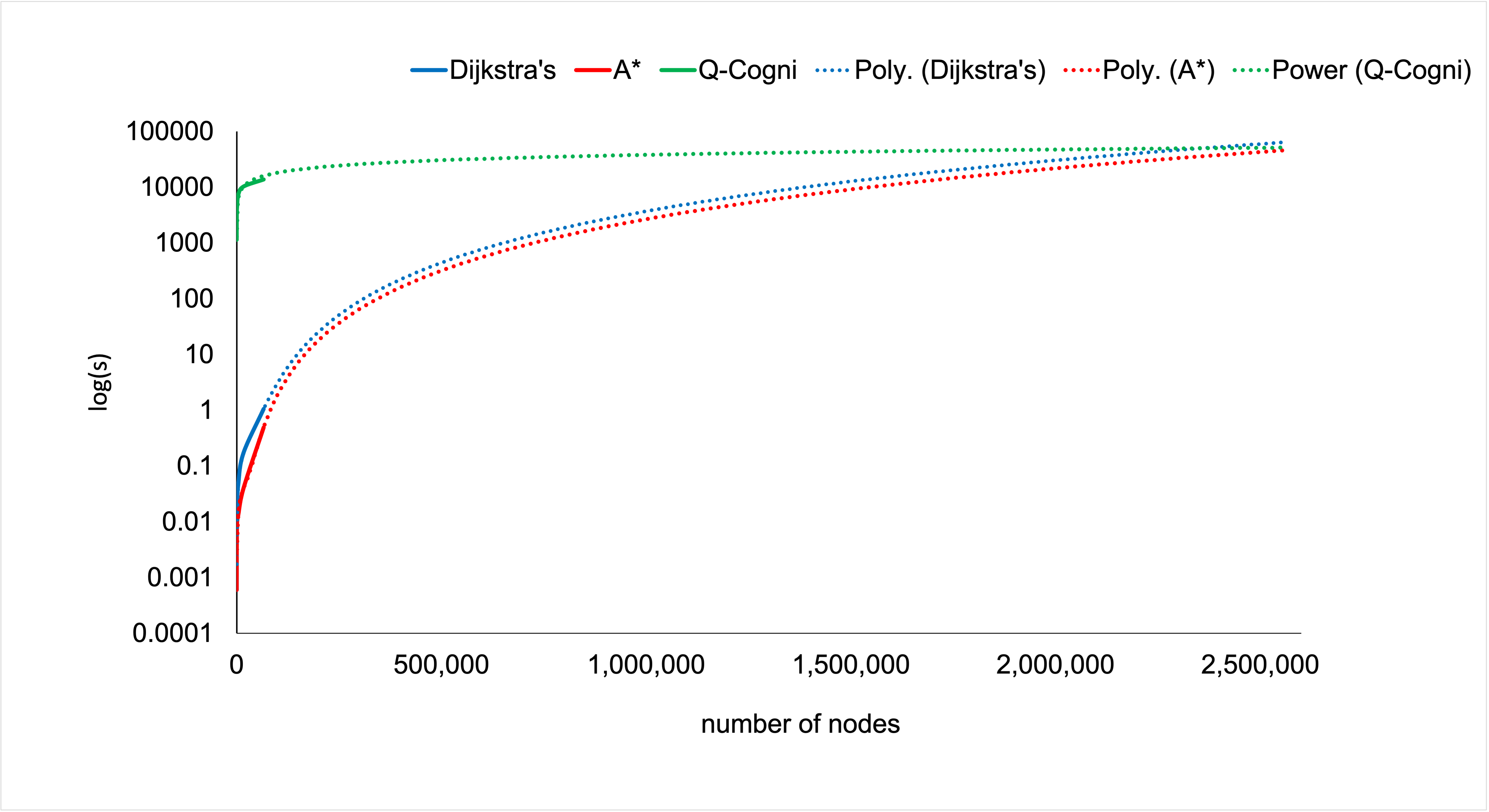}
\caption{Time taken (s) vs. number of nodes for \textit{Taxi-v3} modified environment. Full lines are the experiments performed for each algorithm and dashed lines the extrapolation performed. Log scale.}
\label{Scalability}
\end{figure}

\textbf{Scalability.} \textit{Q-Cogni} excels at large networks. Fig \ref{Scalability} shows the average time taken to identify the optimal tour for varying grid sizes representing a different scale of the \textit{Taxi-v3} environment. We performed experiments for grid sizes from \textit{8$\times$8}, to \textit{512$\times$512} and implemented best-fit curves to extrapolate the increase in problem dimension.

We can observe that \textit{Q-Cogni} takes orders of magnitude longer to identify the optimal tour for a low number of nodes. As the number of nodes increases \textit{Q-Cogni} is much more efficient. This is a product of the sample efficiency delivered within the \textit{Q-Cogni} framework where the causal component enables ``shortcuting" of exploration requirements, thus reducing the need to proportionally increase the observations required by the agent. 

\textbf{A-priori knowledge requirement.} \textit{Q-Cogni} require no prior knowledge of the map. Shortest-path methods require prior knowledge of the graph structure to be effectively applied. For example, in our taxi problem, both Djikstra's and A* require the map upfront. \textit{Q-Cogni} requires the causal structure encoded as a graph, but does not require the map itself. This is a significant advantage to enable application in the real-world as a-priori knowledge can be limited for navigation applications.

\textbf{Transferability.} If the configuration within the map changes (e.g. the initial passenger position), \textit{Q-Cogni} would not need to be retrained. The same agent trained in a configuration of a map can be deployed to another configuration seamlessly. On the other hand, if configuration changes take place, we would require to rerun the shortest-path search algorithms. Therefore, \textit{Q-Cogni} has a significant advantage for dynamic settings, a common characteristic of real-world problems. 

\textbf{Interpretability.} Shortest-path search methods are limited in interpretability of decisions made by the algorithm to derive the optimal tour. The causes in a particular edge is preferred over another are not explicitly described as part of their output. \textit{Q-Cogni} not only is able to provide a full history of reasons in which each decision was made but also the causes and associated probabilities of alternative outcomes at each step. This is another significant advantage on the applicability of \textit{Q-Cogni} for real-world problems in which there is an interface between humans and the agent.

\subsection{Real-World Application: Routing New York City Taxis}
\begin{figure}[t]
\centering
\includegraphics[width=1\columnwidth, height=80pt, clip]{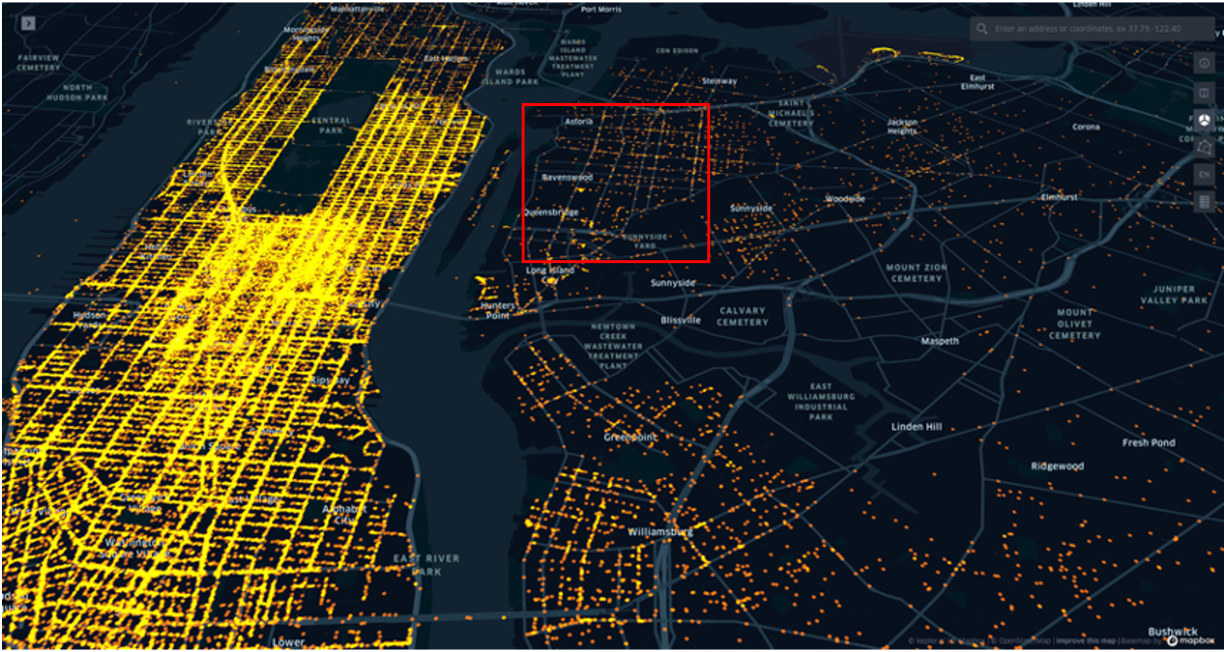}
\caption{Pickup and drop-off points of yellow cabs on New York City 15th October 2022. Highlighted area is the selected for \textit{Q-Cogni} experiments. Built with kepler-gl (https://kepler.gl/).}
\label{NYCmap}
\end{figure}

We use the New York City Taxi \& Limousine Commission trip record data, which contains all taxi movements in New York City from 2013 to date \cite{NYC_data}, to validate the applicability of Q-Cogni in a real-world context. Figure \ref{NYCmap} shows all pickup and drop-off locations of yellow cabs on the 15th of October 2022. We see the highest density of taxi trips being in Manhattan, represented on the left hand side of Figure \ref{NYCmap}. However, we choose the neighborhoods between Long Island City and Astoria, represented in the highlighted area of Figure \ref{NYCmap} as they have a more challenging street configuration than Manhattan.

\begin{figure}[t]
\centering
\includegraphics[width=1\columnwidth, height=120pt, clip]{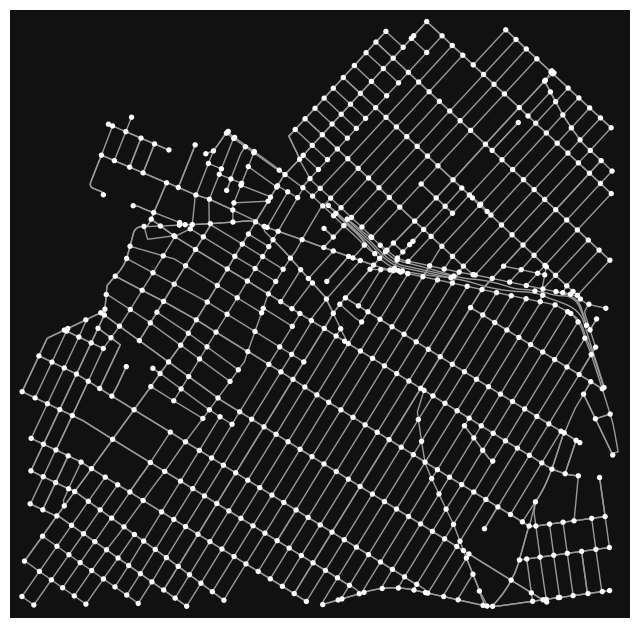}
\caption{Graph representation of the Astoria region in New York City used to train \textit{Q-Cogni}.}
\label{Astoriamap}
\end{figure}

We used the OSMNX library \cite{boeing2017osmnx} to convert the street map into a graph representation where intersections are nodes and edges are streets. We created a custom OpenAI \cite{brockman2016openai} gym environment to enable fitting of the \textit{Bayesian Network} and training of \textit{Q-Cogni}. The resulting graph is shown in Figure \ref{Astoriamap} containing 666 nodes and 1712 edges resulting in a state-action space of size 443,556, a problem $10^3$ larger than our \textit{Taxi-v3} environment.

We use the causal structure derived in Figure \ref{SCMTaxi} and perform a random walk with 1,000,000 steps in the custom built environment to fit a Bayesian Network to the causal model. It is important to appreciate here the transferability and domain agnostic characteristic of our approach, where we leverage on the structure previously discovered for the same task but in a complete different map. 

We train \textit{Q-Cogni} once for 100,000 episodes and evaluate the trained agent against several trips contained in the original dataset without retraining each time the trip configuration changes in the real-world dataset. This is a significant benefit of the approach when comparing to shortest-path search methods. We also compare \textit{Q-Cogni} results against Q-Learning to observe the effects of the causal model against policy quality and compare against Dijkstra's algorithm to observe the effects of policy efficiency.

\begin{figure}[t]
\centering
\includegraphics[width=1\columnwidth, clip]{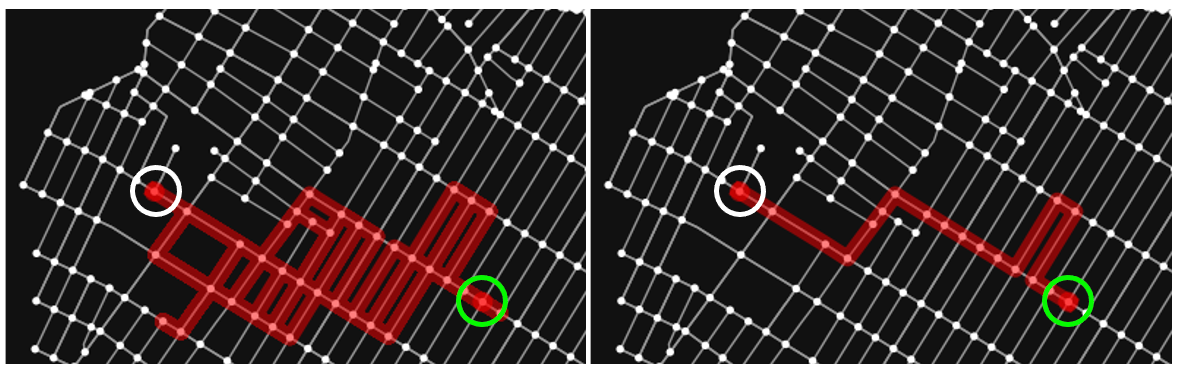}
\caption{Optimal routes generated by Q-Learning (left) and \textit{Q-Cogni} (right) after 100,000 episodes. Origin and destination highlighted with the white and green circles respectively.}
\label{LearningNYC}
\end{figure}

First, Figure \ref{LearningNYC} shows the optimal routes generated by Q-Learning and \textit{Q-Cogni} after 100,000 training episodes for a selected route. We can see that \textit{Q-Cogni} significantly improves the policy generation reaching a near-optimal shortest-path result post training whereas Q-Learning fails to detect the right pickup point and performs multiple loops to get to the destination. Across the 615 trips evaluated, Q-Learning was able to generate only 12\% of routes in which had the same travel distance as \textit{Q-Cogni} generated routes, with the remainder being longer. These results show the benefits of the causal module of \textit{Q-Cogni} towards optimal learning. 

\begin{figure}[t]
\centering
\includegraphics[width=1\columnwidth, clip]{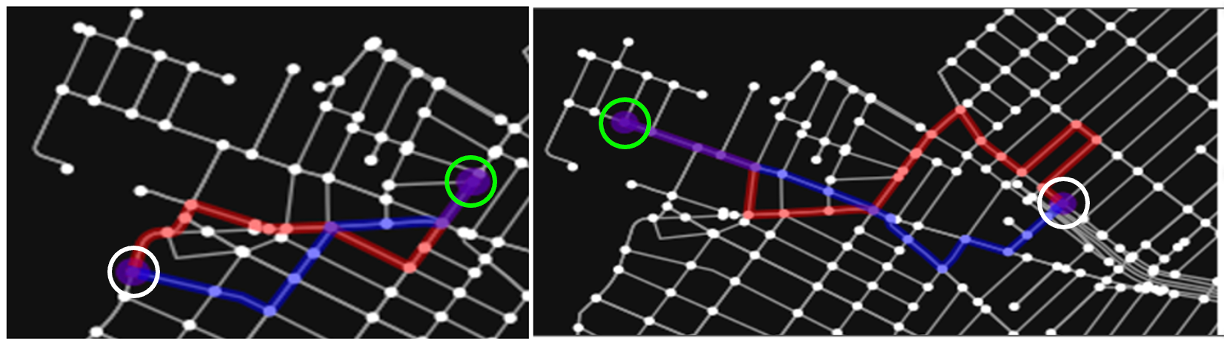}
\caption{\textit{Q-Cogni} (red) generated routes vs. Dijkstra's (blue) for 2 random trips evaluated on the 15th October 2022. Origin and destination highlighted with the white and green circles respectively}
\label{NYCresults}
\end{figure}

In addition, Figure \ref{NYCresults} shows a sample comparison of optimal routes generated with Dijkstra's algorithm and \textit{Q-Cogni}. We observe on the left picture that \textit{Q-Cogni} generates a more efficient route (in red) than Dijkstra's (in blue) measured as the total distance travelled. On the right picture \textit{Q-Cogni} generates a significantly different route which is slightly worse, albeit close in terms of distance. Overall, across the 615 trips evaluated we report 28\% where Q-Cogni generated a shorter route than Dijkstra's, 57\% were the same and 15\% worse.

These results show the applicability of \textit{Q-Cogni} for a real-world case with demonstrated (i) transferability - the causal network obtained in the Taxi-v3 environment can be used for the same task; (ii) no prior knowledge required - \textit{Q-Cogni} does not need to have access to the global map; and (iii) explainability and expandability - where the \textit{Bayesian Network} can be expanded to incorporate other causal relations such as traffic and weather. The application of \textit{Q-Cogni} in this real-world dataset demonstrate a promising framework to bring together causal inference and reinforcement learning to solve relevant and challenging problems.

\section{Conclusion}

We have presented \textit{Q-Cogni}, a novel causal reinforcement learning framework that redesigns \textit{Q-Learning} with an autonomous causal structure discovery method and causal inference as a hybrid model-based and model-free approach. 

We have implemented a framework that leverages upon a data-driven causal structure model discovered autonomously (but flexible enough to accommodate domain knowledge based inputs) and redesigned the \textit{Q-Learning} algorithm to apply causal inference during the learning process in a reinforcement learning setting.

Our approach exploits the causal structural knowledge contained in a reinforcement learning environment to shortcut exploration requirements of a state space by the agent to derive a more robust policy with less training requirements as it increases the sample efficiency of the learning process. Together, these techniques are shown to achieve a superior policy, substantially improve learning efficiency, provide superior interpretability, efficiently scale with higher problem dimensions and more generalisable to varying problem configurations. While these benefits have been illustrated in the context of one specific application – the VRP problem in the navigation domain – it can be applied to any reinforcement learning problem that contains an environment with a implicit representation of the causal relationships between state variables and some level of prior knowledge of the environment dynamics.

We believe that the integration of causality and reinforcement learning will continue to be an attractive area towards human level intelligence for autonomous learning agents. One promising avenue of research is to broaden the integrated approach to continuous state-action spaces such as control environments, a current focus of our research.

\bibliographystyle{named}
\bibliography{ijcai23}

\end{document}